\documentclass[10pt, a4paper, conference]{IEEEtran}

\IEEEoverridecommandlockouts
\usepackage{times,color}
\usepackage{epsfig}
\usepackage{graphicx}
\usepackage{amsmath}
\usepackage{amssymb}

\usepackage{algorithmic}
\usepackage{algorithm}
\usepackage[update,prepend]{epstopdf}

\usepackage{multirow}

\usepackage{caption}
\usepackage{subcaption}

\usepackage{cite}
\ifCLASSINFOpdf
\else
\fi
\begin{document}

\title{Enhanced Random Forest with Image/Patch-Level Learning for Image Understanding}

\author{\IEEEauthorblockN{Wai Lam Hoo\IEEEauthorrefmark{1},
Tae-Kyun Kim\IEEEauthorrefmark{2},
Yuru Pei\IEEEauthorrefmark{3} and
Chee Seng Chan\IEEEauthorrefmark{1}}
\IEEEauthorblockA{\IEEEauthorrefmark{1}Center of Image and Signal Processing, University of Malaya, 50603 Kuala Lumpur, Malaysia\\
Email: wailam88@siswa.um.edu.my; cs.chan@um.edu.my}
\IEEEauthorblockA{\IEEEauthorrefmark{2}Imperial College London, London SW7 2AZ, United Kingdom\\
Email: tk.kim@imperial.ac.uk}
\IEEEauthorblockA{\IEEEauthorrefmark{3}Peking University, Beijing 100871, China\\
Email: peiyuru@cis.pku.edu.cn}}

\maketitle

\begin{abstract}
%\boldmath
Image understanding is an important research domain in the computer vision due to its wide real-world applications. For an image understanding framework that uses the Bag-of-Words model representation, the visual codebook is an essential part. Random forest (RF) as a tree-structure discriminative codebook has been a popular choice.
However, the performance of the RF can be degraded if the local patch labels are poorly assigned. In this paper, we tackle this problem by a novel way to update the RF codebook learning for a more discriminative codebook with the introduction of the soft class labels, estimated from the pLSA model based on a feedback scheme. The feedback scheme is performed on both the image and patch levels respectively, which is in contrast to the state-of-the-art RF codebook learning that focused on either image or patch level only. Experiments on 15-Scene and C-Pascal datasets had shown the effectiveness of the proposed method in image understanding task.
\end{abstract}

\section{Introduction}

Recent studies on image understanding have shown to flavour part-based representation such as the Bag-of-words (BoW) model \cite{Sivic2005,Fei-Fei2005,wailam2013,Lazebnik2006,Moosmann2008,Krapac2011,Niu2012}. Each image given as a set of local patches is represented by a histogram of codewords. Given the codeword representations, topic discovery model such as the pLSA model \cite{Hofmann2001} has been successfully applied to semantic image clustering and unsupervised learning for object categorization and scene understanding. It has been shown that the visual codebook, which is typically obtained by the k-means clustering of the local patches \cite{Sivic2005,Fei-Fei2005,Lazebnik2006}, is a crucial part to achieve good performance.

\begin{figure}[t]
\centering
\begin{subfigure}[t]{0.6\linewidth}
	\includegraphics[height=0.55\linewidth, width=0.95\linewidth]{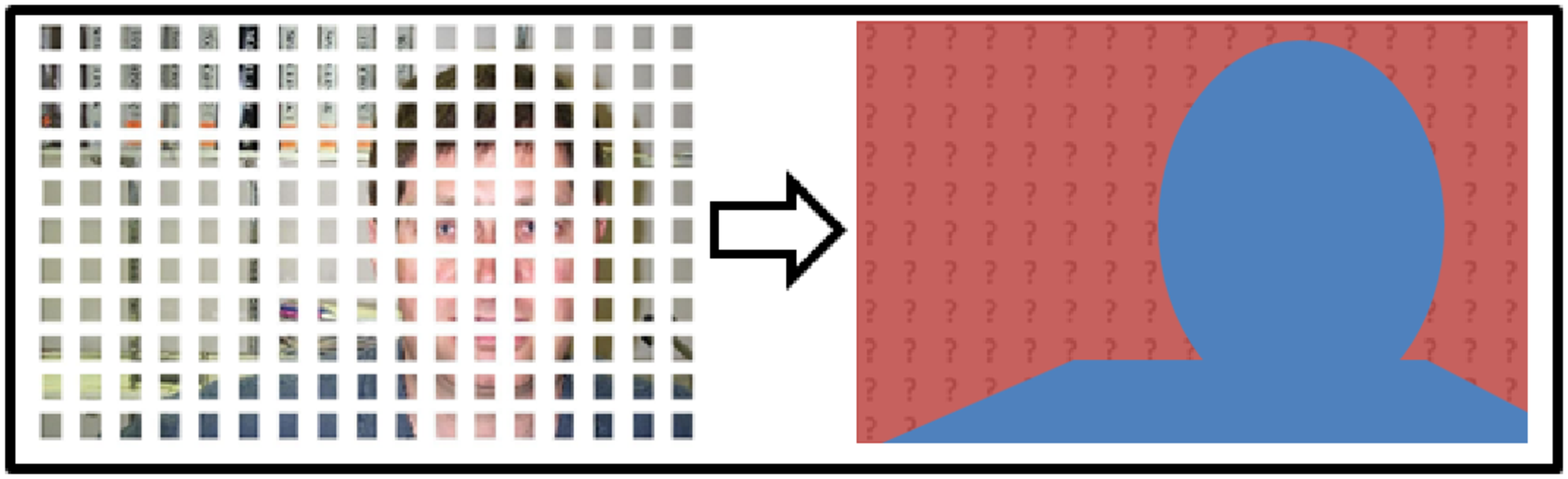}
	\caption{Conventional Approach}
	\label{fig:conventional}
\end{subfigure}
\begin{subfigure}[t]{0.38\linewidth}
	\includegraphics[height=0.75\linewidth, width=0.9\linewidth]{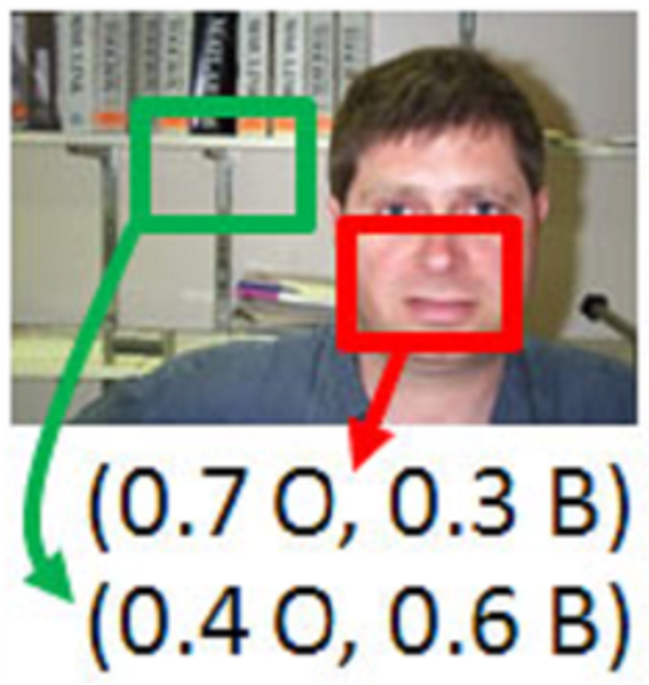}
	\caption{Our Proposed}
	\label{fig:proposed}
\end{subfigure}

\caption{(a) Weakly supervised learning by local patches. It is clear that not all the local patches (i.e. red region) in the image are belong to the "Face" class label. (b) Our proposed method with the use of soft assignment where the local patches are assigned class labels comparatively independently. "O" indicates the object and "B" indicates the background. Best viewed in color.} 
\label{fig:general_idea}
\end{figure}

On the other hand, discriminative codebooks have clearly shown advantage compared to its counterpart (e.g. k-means) when the ground-truth class label of the training image is given. Random Forest (RF) \cite{Krapac2011, Moosmann2008}, an ensemble of decision trees with randomization, appears to be a very fast algorithm compared to the k-means codebook where the clustering and vector quantization process are highly time-demanding. Besides, the RF also shows its discriminative power as an effective codebook for object categorization and segmentation. However, this advantage is heavily relied on the accuracy of the ground-truth class label. For example, let us assume that the ground-truth class label in Figure \ref{fig:conventional} is belong to the `Face' class, and hence all the local patches for the image will be associated with the `Face' class label. However, we can clearly notice that  not all the local patches are belong to the `Face' class label (e.g. the background patches (red region) should not belong to the `Face' class label). As such, during training the RF codebook, the background patches which are wrongly labeled in this case will greatly degrade the discriminative power of the RF codebook.

\begin{figure*}[t]
\centering
\includegraphics[height=0.3\linewidth, width=0.85\linewidth]%[height=6cm,width=12cm]
{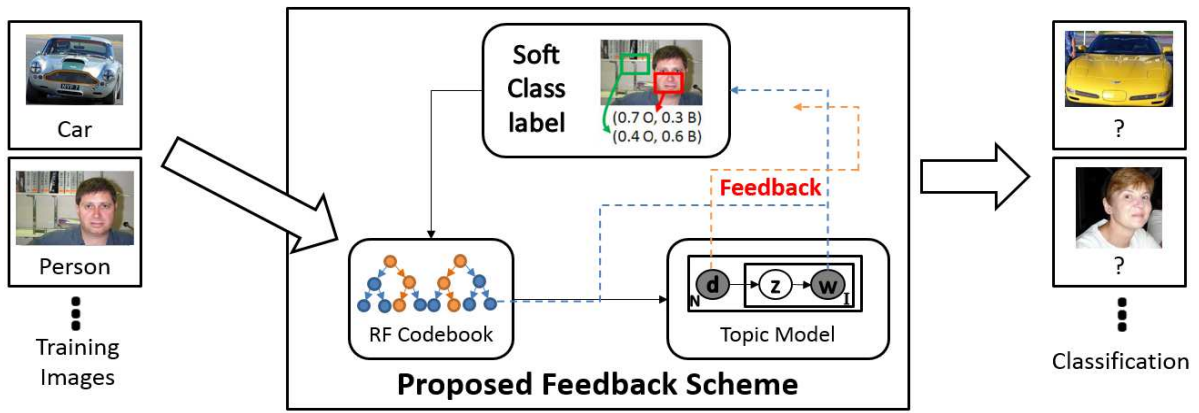}
\caption{An overview of the proposed feedback framework. The blue dotted line indicates patch information from the RF and pLSA, while orange dotted line indicates the image level information from the pLSA. Best viewed in color.}
\label{fig:framework}
\end{figure*}
%++++++++++++++++++++++++++++++

In this paper, we present a novel way of learning the RF by introducing the soft class labels, estimated from the pLSA model based on a feedback scheme. Particularly, we adapt the RF node split strategy to account for the soft class labels obtained from the initial weak pLSA classifier.  The feedback scheme can be performed on the image and patch levels respectively; and we anticipated that the RF re-learning and the pLSA re-training in the close feedback loop will improve the discriminative power of the codebooks. Experiments on 15-Scene and C-Pascal datasets have shown that the proposed codebook outperforms the state-of-the-art methods such as \cite{Krapac2011, Moosmann2008}.  

This rest of the paper is arranged as follows: Section \ref{Sec2} discusses the recent developments in related topics including the visual codebook learning and topic model. Section \ref{ref:Sec3} details the proposed framework. We show the experimental results in Section \ref{r:Sec4}. Finally, discussions and conclusion are drawn in Section \ref{r:Sec5}-\ref{r:Sec6}, respectively.
%****************************************************************************

%============================================================================
% Related Work
%============================================================================
\section{Related Work}
\label{Sec2}
Visual codebook learning is an essential pipeline in the BoW representation. In order to find the optimal codewords, unsupervised methods such as the k-means \cite{Sivic2005} and kd-tree \cite{muja2009fast} had been employed. However, recent research work focused on learning the visual codebook using labeled images (i.e. supervised manner), in order to have a better discrimination on the codebook learning.  

According to \cite{Breiman2001}, RF which offers discriminative characteristics compared to the generative approaches such as the K-means is a popular choice. For instance, Moosmann et al. \cite{Moosmann2008} used the image classification results from the RF as a feedback mechanism in the interest point detector to create a saliency map. The object location from the saliency map was then used in re-learning the RF for a better codebook. We denote this approach as the patch-level feedback scheme. Krapac et al.\cite{Krapac2011} proposed another variant of feedback scheme in the RF re-learning. They performed codebook learning by alternating the quantizers and classifiers to maximize the classification performance. In each split node, the data set is separated into the training and validating set to evaluate the classifier. Finally, the results from the classifier are feedback to the nodes for the optimal node splitting. Similar joint learning approaches between the dictionary and the classifier were also studied in \cite{Boureau2010,Yang2010,Lian2010} and we denote these approaches as the image-level feedback scheme. However, in this paper, a special feedback scheme that utilizes both the image and patch-level is proposed. In particular, we utilize the topic model for the soft assignment on image and patch labels to further improve the RF learning for a more discriminative codebook. 

Topic models are widely applied in image classification \cite{Quelhas2007}. The topic models are particularly effective when pairing with the BoW representation, where the models group ambiguous codewords together and generate a topic distribution over a codebook. One of the most popular topic model is the pLSA \cite{Hofmann2001} which serves as a mid-level clustering method and tries to find the relationship of codewords. The codewords are grouped together for some meaningful representations. For instance, a "Face" class image as illustrated in Figure \ref{fig:conventional} where ideally the pLSA will cluster the image regions into two parts. One part represents the face by grouping eyes, hairs and mouths; while the other represents the background (e.g. book, sky) by codewords. Similar to other clustering methods, the pLSA also requires to find the optimal number of topics to represent a particular image effectively. In this paper, we employ the pLSA to estimate the soft class label for the training data. In turn, these soft class labels are combined together to re-learn (enhance) the RF codebook and follow by re-train the pLSA model in a novel feedback scheme. 

In summary, our main contribution is the introduction of \textbf{a special feedback scheme where the soft class labels} from a topic model are used to update the RF learning for a more discriminative codebook. The feedback scheme can be performed on the image and patch levels respectively for the image understanding task. Such a framework is more superior compare to the conventional solutions \cite{Moosmann2008,Krapac2011} as it includes both the image-level information, estimated from the topic model, as well as the patch-level information, estimated from both the topic model and RF codebook respectively. This is in contrary to \cite{Moosmann2008} that only employs the patch-level information, and \cite{Krapac2011} that only employs the image-level information.

%============================================================================
% Methodology
%============================================================================

\section{Methodology}
\label{ref:Sec3}
%=============================================================================================

%=============================================================================================

The proposed method is illustrated in Figure \ref{fig:framework}. First, we learn a weak RF using the local patches that are associated with the ground-truth image class label. Treating the RF leafnodes as codewords, we build the BoW representation from the RF codebook. Secondly, we train a weak pLSA model from the RF codebook in order to estimate the soft class labels. Thirdly, a feedback scheme where the soft class labels from the weak topic model are used to update the RF (re-learning), and follow by a new (enhanced) pLSA model is trained from the refined RF codebook. The feedback scheme will iterate until the convergence criteria is satisfied. Finally, classification is performed using the converged pLSA model.

%=============================================================================================

\subsection{Initial RF codebook learning and pLSA model training}
\label{sec:init_learning}

We start with a weak classifier construction by a RF codebook and a pLSA model from a set of labeled training images. The RF as an ensemble of the random decision trees by bagging provides a very fast way of codebook learning and quantization. Moreover, when the class labels are available, it has an advantage as a discriminative codebook. The random decision tree is constructed using a random subset of the training data with replacement. The labeled training images $I'_{\mathbf{N}}=\{x_i,l_i\}$ at a specific node $\mathbf{N}$, where ${x_i}$, ${l_i}$ are the feature vectors of the local patches and the corresponding class labels respectively, are recursively splitted into left  $I'_{\mathcal{L}}$ and right $I'_{\mathcal{R}}$ subsets, according to a set of thresholds $T$ and a split function $f$, as
%=============================================================================================
\begin{equation}
I'_{\mathcal{L}} = \{x_i \subset I'_{\mathbf{N}} | f(x_i) < T_t\}, I'_{\mathcal{R}} = I'_{\mathbf{N}} \setminus I'_{\mathcal{L}}.
\end{equation}
%=============================================================================================

At each split node, a random subset of the features are generated and compare to $T$. The ones that maximize the expected information gain $\triangle E$ are selected. Specifically, at each split node:
%=============================================================================================
\begin{equation}
\triangle E = E(I'_{\mathbf{N}}) -\sum_{i=\mathcal{L},\mathcal{R}} \frac{\mid I'_i \mid}{ \mid I'_{\mathbf{N}} \mid}E(I'_i),
\end{equation}
%=============================================================================================
\begin{equation}
E(I'_i) = p(l_i) \enspace (log_2 \enspace p(l_i)),
\label{eq:SE}
\end{equation}
%=============================================================================================
where $E(I'_i)$ is the Shannon Entropy of the probability class histogram $p(l_i)$ of the training images $I$. The leafnodes of all the trees in the forest will serves as a codebook. Then, the feature vectors ${x_i}$ are quantized by the learnt RF codebook to form the BoW representation. Initial pLSA model is trained from the BoW whose element $\{w_j,d_n\}$ stores the number of occurrences of a codeword $w_j$ in the image $d_n$, where $j$ is number of codewords and $n$ is the number of images. The image topics $z_k$ are selected accordingly to the image-specific topic distribution $p(z_k|d_n)$, where $k$ is number of topics. These parameters are estimated by maximizing the log-likelihood algorithm as:
%=============================================================================================
\begin{equation}
p(d_n,w_j) = p(d_n) \sum_{k=1}^{K} p(w_j|z_k) p(z_k|d_n),
\end{equation}
%=============================================================================================
\noindent and we can estimate the image-specific topic distribution $P(z_k|d_n)$ by
%=============================================================================================
\begin{equation}
p(w_j|d_n) = \sum_{k=1}^{K} p(w_j|z_k) p(z_k|d_n),
\end{equation}
%=============================================================================================

\subsection{Soft class labels}
\label{sec:soft_class_labels}

In this paper, we introduce a soft class label in which the local patches are assigned class labels comparatively independently. As to one image, the local patches in the object regions are assigned the labels with respect to the object classes, while the local patches in the background regions are assigned to background labels. The feedback in the image and patch level assigns confidences to different class labels to every single patch, which produce discriminative label to patches in different regions. This is in contrast to the conventional solutions \cite{Krapac2011, Moosmann2008} where all the local patches in the same image are assigned the same image label. The confusions can be caused in the codebook learning when the local patches actually in the background are assigned labels of objects as depicted in Figure \ref{fig:conventional}. 

The soft class labels are estimated from $p(z_k|d_n)$ (i.e. from the initial pLSA model), and hence we can calculate the image-codeword-specific topic distribution $p(z_k|w_j,d_n)$ as:

%%=============================================================================================
\begin{equation}
p(z_k|w_j,d_n) =
\frac{p(w_j|z_k)p(z_k|d_n)}{\sum_{k=1}^{K} p(w_j|z_k)p(z_k|d_n)}.
\end{equation}
%=============================================================================================

In here, it is assumed that $p(z_k|d_n)$ has a close relationship to the image-level soft class label $p(c_m|d_n)$, and so we can estimate $p(c_m|d_n)$ from the available labeled training images where $c$ is the object class, and $m$ is the number of classes.  Concretely, we define a Dominant Topic representation $\mathcal{T}_m$ for each $c_m$, i.e. a set of $z_k$ that are representative for a particular class $c_m$. So, we can calculate class-specific topic distribution $p(z_k|d_m)$ as:
%%=============================================================================================
\begin{equation}
p(z_k|d_m) = \frac{\sum_{n \subset m} p(z_k|d_n)}{\sum_{m=1}^M p(z_k|d_m)}.
\end{equation}

Then, for $z_k$ that satisfy the condition $p(z_k|d_m) > 1/K$, we assign to $\mathcal{T}_m$ and compute $p(c_m|d_n)$:

\begin{equation}
p(c_m|d_n) = \frac{\sum_{k \subset\mathcal{T}_m} p(z_k|d_n)}{\sum_{m=1}^{M} p(c_m|d_n)}. 
\label{mapping}
\end{equation}

However, every local patch has different probability values based on the relationship between the codewords $w_j$ and patch feature vectors $x_i$. During the quantization process, $x_i$ is represented by $J$ codewords, where $J = R \times S$ and $R$ is the total number of trees used in codebook learning, while $S$ is the total leafnodes per tree. Conventionally, each $w_i$ gives a class probability based on the local patches $p(c|x_i)$. However, we treat each $w_i$ as an individual `class', and so $p(c|x_i)$ can be rewritten as codeword probability of the feature vector, $p(w_j|x_i)$:
%=============================================================================================

\begin{equation}
p(w_j|x_i) = \frac{1}{J} \sum_{r=1}^{R} \sum_{s=1}^{S} p(w_{r,s}|x_i).
\end{equation}
%=============================================================================================
The image-patch-specific topic distributions $p(z_k|x_i,d_n)$ are then determined by summing the corresponding $p(w_j|x_i)$:
%=============================================================================================

\begin{equation}
p(z_k|x_i,d_n) = \frac{p(z_k|w_j,d_n)p(w_j|x_i)}{\sum_{k=1}^K p(z_k|w_j,d_n)p(w_j|x_i)}.
\end{equation}
%=============================================================================================
In order to estimate the soft class labels for each local patches $p(c_m|x_i,d_n)$, we utilize $p(z_k|x_i,d_n)$ and $\mathcal{T}_m$. As before, the $\mathcal{T}_m$ is a representation that is defined based on $p(z_k|d_m)$ (i.e. a set of $z_k$ that significantly represents $c_m$), hence all $x_i$ that belong to the same $c_m$ will share the same $\mathcal{T}_m$. With this, we can compute $p(c_m|x_i,d_n)$ as:  
%=============================================================================================

\begin{equation}
p(c_m|x_i,d_n) = \frac{\sum_{k \subset\mathcal{T}_m} p(z_k|x_i,d_n)}{\sum_{m=1}^M p(c_m|x_i,d_n)}.
\label{eee1}
\end{equation}

%=============================================================================================
%=============================================================================================

\subsection{Feedback mechanism}% Codebook Re-learning and topic model Re-Training}
%=============================================================================================

In order to re-learn (enhance) the weak RF codebook, we use the soft class label estimated from the topic model (Section \ref{sec:soft_class_labels}). Since $x_i$ labels change from hard assignment $p(l_i)$ to soft assignment $p(c_m|x_i,d_n)$, we refine the spliting criterion - Shannon Entropy $E$ in Eq. \ref{eq:SE} and we compute the new probability class histogram $p'(l_i)$ as:

%=============================================================================================

\begin{equation}
p'(l_i) = \frac{ p(c_m|x_i,d_n)}{\sum_{m=1}^M p(c_m|x_i,d_n)}.
\end{equation}
%=============================================================================================

\noindent Other settings of the RF codebook is remain the same. Then, an enhanced pLSA model is trained from the new BoW based on the refined RF codebook.

\subsection{Classification}
\label{sec:classification}

Finally, we estimate the image-specific topic distribution of the test images $p(z_k|d_{\text{test}})$ as: 
 
%=============================================================================================

\begin{equation}
p({w_j}|{d_{\text{test}}}) = \sum\limits_{k = 1}^K {p({w_j}|{z_k})p({z_k}|{d_{\text{test}}})} .
\end{equation}
%=============================================================================================
In order to estimate the image-level soft class label of test image $p(c_m|d_{\text{test}})$, we use $p({z_k}|{d_{\text{test}}})$ and similar $\mathcal{T}_m$ from the Section \ref{sec:soft_class_labels}:
%based on a normalization process that similar to Eq. \ref{eee1}. 
%=======================================================
\begin{equation}
p(c_m|d_{\text{test}}) = \frac{\sum_{k \subset \mathcal{T}_m} p(z_k|d_{\text{test}})}{\sum_{m=1}^{M} p(c_m|d_{\text{test}})} .
\label{eee}
\end{equation}
%=======================================================\\
A set of class-specific thresholds, $h_m$ are identified from the soft class labels of a set of training images $p(c_m|d_{\text{train}})$. For images that $p(c_m|d_{\text{test}}) > h_m$, it will be denoted as the correct classification. Algorithm \ref{Algo:Modeling} summarizes the proposed method. 
%=============================================================================================

\begin{figure}[t]
\centering
\includegraphics[height=0.68\linewidth, width=0.95\linewidth]{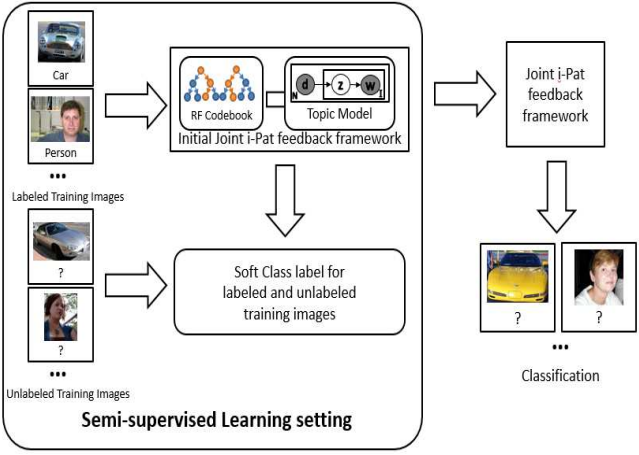}
\caption{Semi-supervised learning for the proposed framework.%and domain adaptation framework
}
\label{fig:SSL}
\end{figure}

\begin{algorithm}[t]
\caption{Proposed Framework}
\begin{algorithmic} 
\REQUIRE A set of labeled training image patches $\{x_i,l_i\}$.
\ENSURE All parameters are set: total number of trees $R$, total number of leafnodes $E$ and total number of topics, $K$ 
\STATE 1. Initial learning of the RF using $\{x_i,l_i\}$.
\STATE 2. Initial training of the pLSA model using the BoW histogram based on the initial RF as in step 1.

\REPEAT
\STATE a. Infer soft class labels $p(c_m|x_i,d_n)$
to associate with $x_i$.\\
b. Re-learn RF using $x_i$ that associated with corresponding $p(c_m|x_i,d_n)$.\\
c. Re-train the pLSA model based on the initial RF as in step 2b.
\UNTIL{Convergence criteria is satisfied}
\STATE 3. Classification using the final pLSA model.
\end{algorithmic}
\label{Algo:Modeling}
\end{algorithm}

%=============================================================================================
\subsection{Semi-supervised Learning}

In a real world, however, few data are labeled and obtaining exhaustive annotation is impractically expensive. As such, the semi-supervised methods have been studied. In order to show the capability of the proposed method when confronted with with the unlabeled data in the training image set, we show here how the proposed framework can be extended to the semi-supervised learning paradigm. Conventionally, the RF cannot deal with semi-supervised learning since it needs image class labels for each features associated with it when training. Inspired by \cite{Leistner2009}, we discover a possibility to extend the proposed framework to semi-supervised method. Refer to Figure \ref{fig:SSL}, all the operations are similar as to Algorithm \ref{Algo:Modeling} except we will estimate the soft class labels for both the labeled and unlabeled training images. %Refer to Figure \ref{fig:SSL}, for a set of training images that contains labeled and unlabeled images, we first learn the weak RF codebook and train the weak pLSA model from the labeled training images only. Secondly, we estimate the soft class labels for both the labeled and unlabeled training images. Then, the rest of the process is similar as to the proposed framework. 
%=============================================================================================

%============================================================================
% Experimental Result
%============================================================================
\section{Experiments}
\label{r:Sec4}

In the experiment, we employed 15-Scene and C-Pascal datasets to test the effectiveness of the proposed framework, and as a comparison to the state-of-the-art methods. The 15-Scene dataset \cite{Lazebnik2006} consists of both indoor and outdoor scene images. Each class consists of 200-400 images with 300 $\times$ 250 pixels respectively. The C-Pascal dataset \cite{ebert2011pick} is created based on the bounding box annotations for each object class from the PASCAL VOC challenge 2008 dataset \cite{pascal-voc-2008}. As such, the classification can be evaluated in a multi-classes setting. This dataset contains 4450 images from 20 object classes with varying object poses and background clutters.

For both datasets, we perform dense SIFT on a patch size $=$ 8 and step size $=$ 4. We choose a small patch size and step size as some of the images in the C-Pascal dataset are low resolution. Beside that, for any image with edge $>$ 300 pixels, it will be resized to 300 pixel but the aspect ratio is retained. For the RF codebook settings, we use 10 random trees with 100 leafnodes, resulting in 1000 codeword histogram. For learning the pLSA, we use 20 topics and 100 training images for the 15-Scene dataset, while 30 training images for the C-Pascal dataset.

\noindent \textbf{Experimental result:} For the 15-Scene dataset results that depicted in Table \ref{table:result_15scene}, it is noticed that our proposed method outperform the state-of-the-art method ScSPM  \cite{yang2009linear} with an improvement of 2.2\%. Though the improvement seems narrow, one must note that the employed ScSPM is in the optimum settings as published in their paper. 
To have a better understanding of the performance of the ScSPM and our proposed solution, we reimplemented the ScSPM into two different configurations: (ScSPM$_a$) - a 1024 bases with no Spatial Pyramid Matching (SPM), and (ScSPM$_b$) - a 64 bases with 3-level SPM. Both configurations will result in 1000 bases/codewords that is similar to our proposed framework for a fair comparison.
Again, the proposed method outperforms the ScSPM$_a$ and ScSPM$_b$ with improvement of 19.35\% and 10.49\%, respectively. Compare to the ERC-forest \cite{Moosmann2008} which is a patch-level feedback scheme solution, we also perform well as we have 8.64\% improvement. This has shown that the ERC forest is affected by the wrongly labeled local patches. 

Although we are slightly weaker (i.e. a very small margin of 1.12\%) to the Tree Quantizer \cite{Krapac2011} when labeled training image = 100, in our experiment, we used a simple dense SIFT feature compared to their work, which sampled features on the original image scale, as well as at four down-sampled versions (i.e. each time, the image is re-scaled by a factor of 1.2). Also, in their work, they used 15 (class) $\times$ 10 (trees) $\times$ 100 (leafnodes) for their codebook representation in the 15-Scene dataset experiment, which is a lot larger to ours, and thus resulting in much higher computational cost compared to our proposed method..

%=============================================================================================
\begin{table}[t]
\caption{Accuracy (in terms of \%) on the 15-Scene Dataset and comparison to the state-of-the-art methods}
\begin{center}
\resizebox{0.85\linewidth}{!}{
\begin{tabular}{|c|c|c|}
\hline
Labeled training image & 10 & 100\\
\hline
Total training image & \multicolumn{2}{|c|}{100}\\
\hline
\hline
ERC Forest \cite{Moosmann2008} & 49.95 & 73.84 % 83.9 $\pm$ 0.6 
\\
Tree Quantizer \cite{Krapac2011} & 48.14 & \textbf{83.60} \\
\hline
Proposed method & \textbf{77.38} & 82.48\\
\hline
\hline
KSPM \cite{Lazebnik2006} & \multicolumn{2}{|c|}{81.40} \\
KC \cite{Gemert2010} & \multicolumn{2}{|c|}{76.67} \\
ScSPM \cite{yang2009linear} & \multicolumn{2}{|c|}{80.28} \\
ScSPM, base 1024, no SPM & \multicolumn{2}{|c|}{63.13}\\
ScSPM, base 64, 3 level SPM & \multicolumn{2}{|c|}{71.99}\\ 
\hline
\end{tabular}
}
\end{center}
\label{table:result_15scene}
\end{table}
%=============================================================================================
\begin{table}
\caption{Accuracy (in terms of \%) on the C-Pascal dataset and comparison to the state-of-the-art methods}
\begin{center}
\resizebox{0.9\linewidth}{!}{
\begin{tabular}{|c|c|}
\hline
Algorithm & Accuracy (\%) \\
\hline
Multiple features + rank \cite{ebert2010extracting} & 45.50 \\
LP+ITML (best case) \cite{ebert2011pick} (5 training) & 36.40 \\
NN with active learning using spDSIFT \cite{ebert2013semi} & 32.90 \\
RALF \cite{ebert2012ralf} & 37.30 \\
GP-OA-Var\cite{bodesheim2013approximations} (area under AUC) & 76.26 \\
\hline
Proposed method (30 training, 5 labeled) & 75.81\\
Proposed method (30 training, 30 labeled) & \textbf{85.29}\\
\hline
\end{tabular}
}
\end{center}
\label{table:result_CPascal}
\end{table}

%=============================================================================================
\begin{table}[t]
\caption{Convergence analysis (accuracy) for 15-Scene and C-Pascal datasets in different training settings.}
\begin{center}
\resizebox{0.95\linewidth}{!}{
\begin{tabular}{|c|c||c|c|c|c|}
\hline 
\multirow{2}{*}{Dataset}  & Number of & Initial   & 1st  & \multirow{2}{*}{Convergence} & Best \\ 
 &  Labeled Images &  Learning  &  Iteration &  &  Result \\
\hline
\hline
15 Scene  & 10 & 76.19\% & 76.63\% & \textbf{77.38\%} & \textbf{77.38\%} \\
\hline
15 Scene  & 100 & 76.10\% & 81.45\% & \textbf{82.48\%} & \textbf{82.48\%} \\
\hline
\hline
CPascal &  5 & 72.51\% & 75.56\% & \textbf{75.81\%} & \textbf{75.81\%}\\
\hline
CPascal &  30 & 67.81\% & \textbf{85.20\%} & 84.77\% & \textbf{85.20\%} \\
\hline
\end{tabular}
}
\end{center}
\label{table:convergence_15scene}
\end{table}
%=============================================================================================

\begin{figure}[t]
\centering
\includegraphics[height=0.5\linewidth, width=0.9\linewidth]{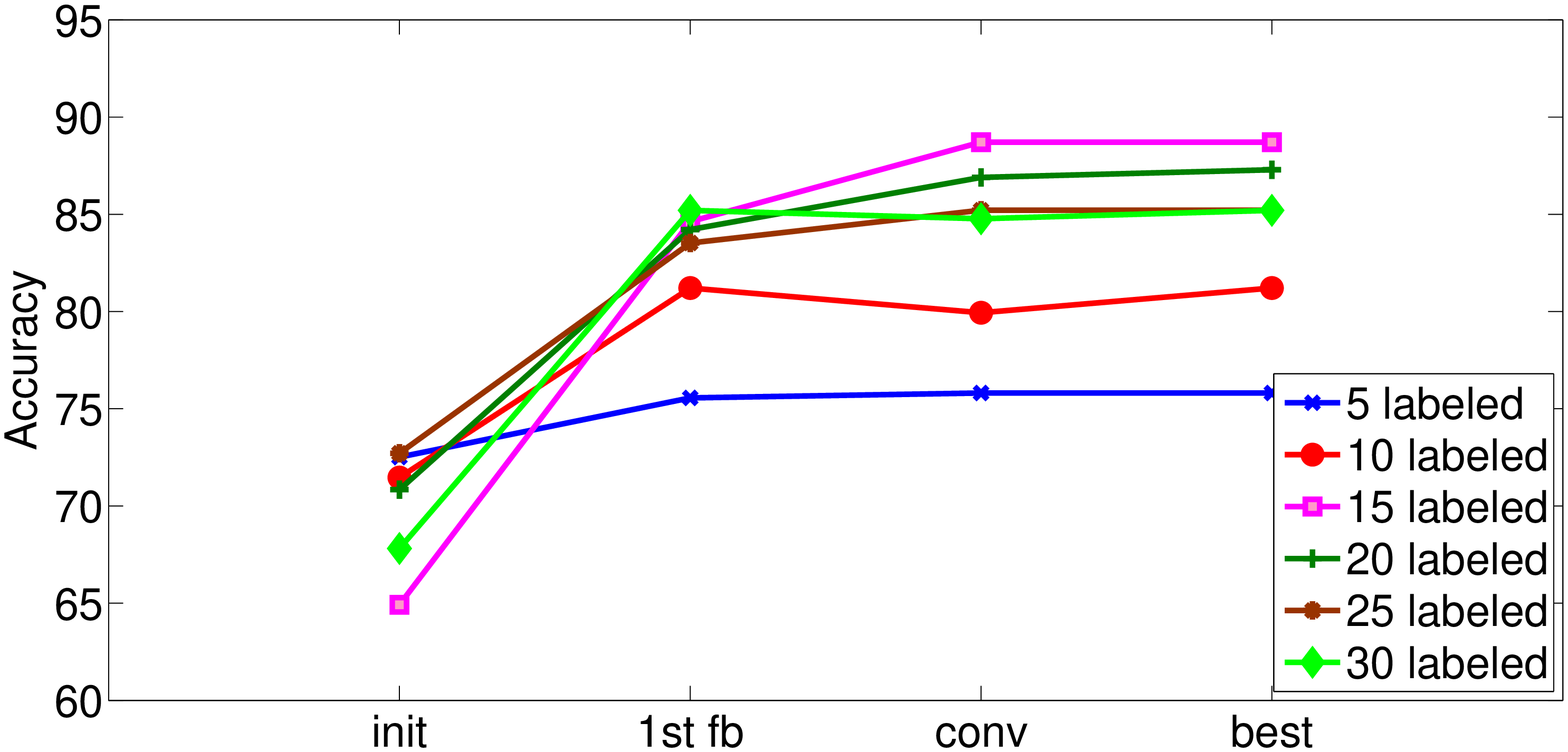} % fig1
\caption{Analysis on C-Pascal dataset, specifically the convergence analysis in semi-supervised learning. init: Result by initial pLSA; 1st fb: Result for first feedback; conv: Convergence result; best: Best result achieved out of all iteration. Best viewed in color.}
\label{fig:conv_ssl_cpascal}
\end{figure}

For the C-Pascal dataset results show in Table \ref{table:result_CPascal}, our proposed method yet again outperform the conventional solutions \cite{bodesheim2013approximations, ebert2010extracting, ebert2011pick, ebert2013semi, ebert2012ralf}. Even with a lower amount of labeled training images (i.e. only 5 labeled in 30 training images), the proposed method still capable to achieve comparable performance (i.e. rank 2 overall) with an accuracy of 75.81\%.

\noindent \textbf{Convergence (comparison to conventional pLSA).} Based on Table \ref{table:convergence_15scene}, we treat each stage of the proposed feedback scheme as an independent pLSA classifier, and report the results in iteration basis. Results of the first feedback for both methods tend to be very close to the final converged results. 
This is expected because the first feedback and the following feedbacks (iteration) have the same amount of features and soft class labels to learn the RF codebook, comparing to weak pLSA model which is build based on a limited number of labeled training images. 
Therefore the improvement on the initial learning and after the first feedback is more significant.  Also, the final convergence result doesn't necessary to be the best classification model, e.g. in Figure \ref{fig:conv_ssl_cpascal}, the final convergence models in the C-Pascal dataset that employed 10, 20 and 30 labeled training images respectively, are not the best classification model.

\noindent \textbf{Semi-supervised learning:} In Table \ref{table:result_15scene}, for the 15-Scene dataset, our SSL settings (i.e. 10\% of the total training image as labeled training image) have comparable result to the state-of-the-art solutions despite limited labeled training images are available. Note that this amount of labeled training images are very limited, and very close to unsupervised learning using the ScSPM.
Our proposed method are weaker to both the KSPM and ScSPM for 4.02\% and 2.90\% respectively, but we have an improvement of 14.25\% and 5.39\% compare to ScSPM$_a$ and ScSPM$_b$ respectively. This shows the flexibility and effectiveness of our proposed method working in the SSL environment. In the meantime, the ERC forest and Tree Quantizer methods degrade drastically to around 50\% accuracy in this SSL environment, because both methods can only utilized the labeled training images during the RF learning. Therefore, the number of images used during the RF learning is very limited, and results in poor performance. 
Bear in mind that our classifier is based on the pLSA topic model, which is a generative approach. Therefore, we believe that the classification result can be  better if a hybrid approach as in \cite{Bosch2008} is applied. 

The C-Pascal dataset result is explained in Figure \ref{fig:conv_ssl_cpascal}, where the experiments are conducted with the amount of labeled training images increase gradually by 5. The experiments clearly show the improvement from the feedback mechanism for various settings. 
Also, fully labeled settings doesn't necessary provide the best result because there will be more local patches that are wrongly labeled in the initial learning, which weaken the initial RF. 
With considerable amount of unlabeled training images (i.e. in the experiment, roughly half of from the total training images), the RF have high chances to be enhanced by the unlabeled training images with the feedback mechanism, and able to achieve better results.

\begin{figure}[t]
\centering
\begin{subfigure}[b]{0.15\textwidth}
	\centering
	\includegraphics[height=0.5\textwidth, width=1\textwidth]{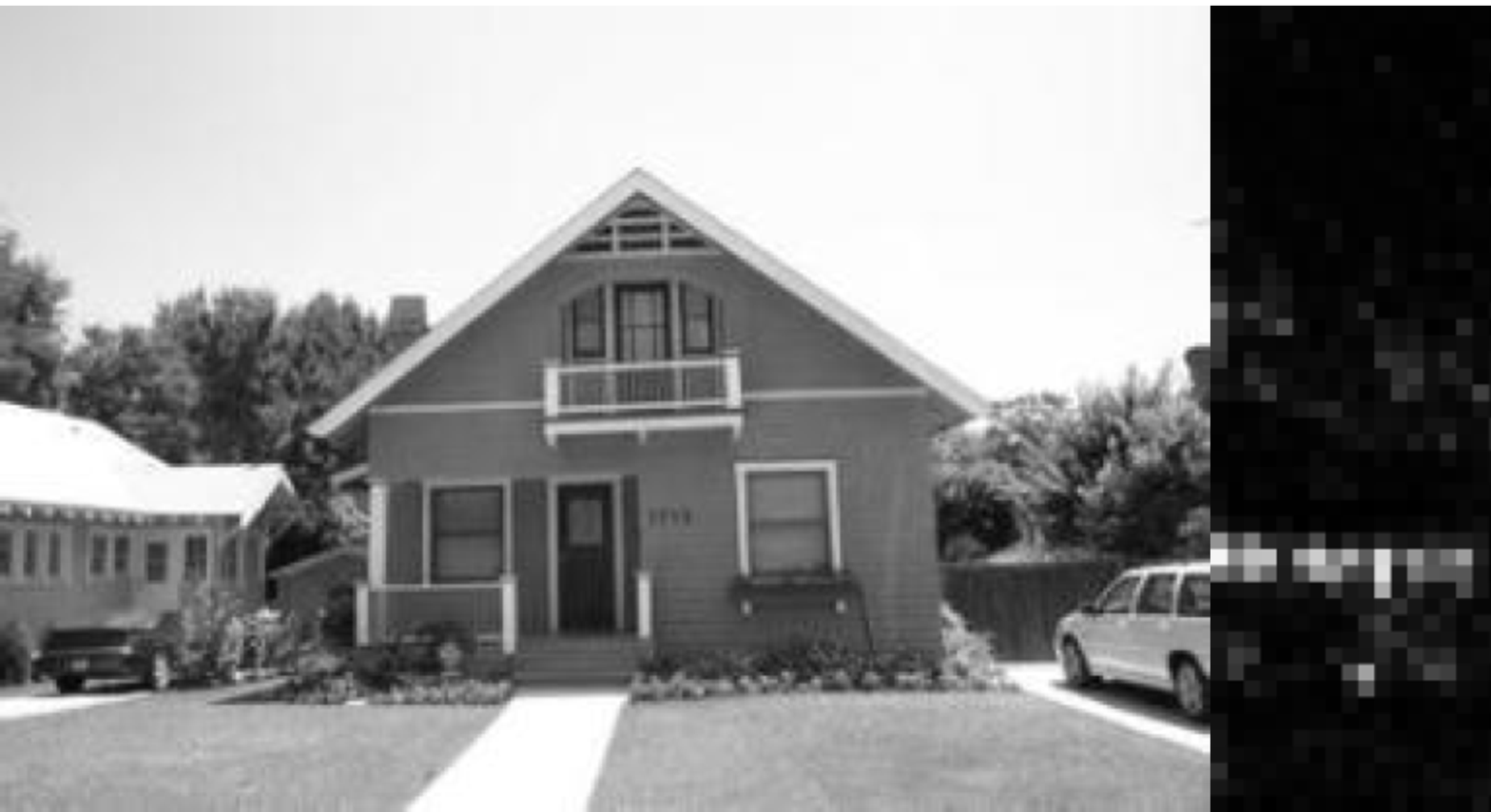} \\ 
	\caption{CALsuburb}	
\end{subfigure}
\begin{subfigure}[b]{0.15\textwidth}
	\centering
	\includegraphics[height=0.5\textwidth, width=1\textwidth]{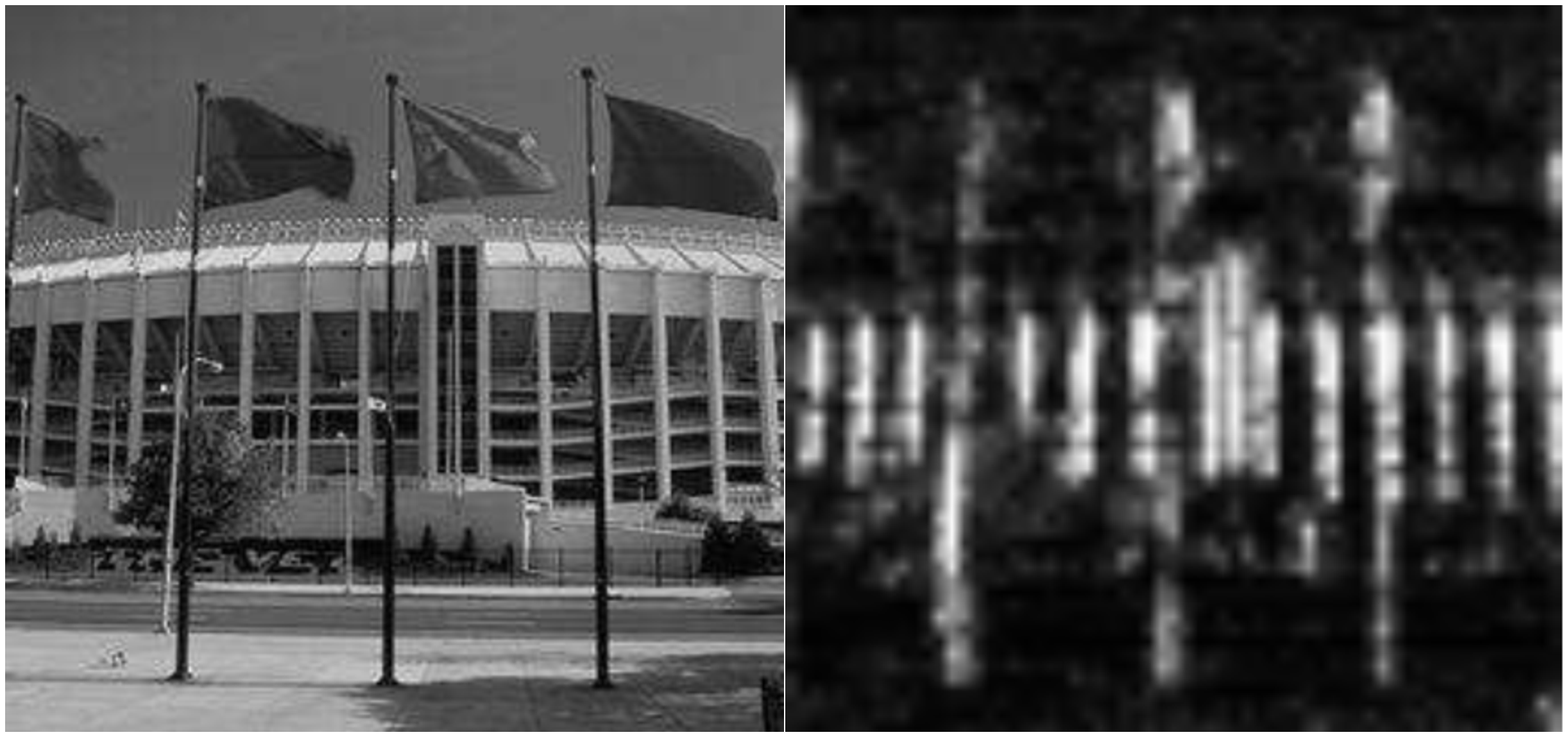} \\ 
	\caption{Inside city}	
\end{subfigure}
\begin{subfigure}[b]{0.15\textwidth}
	\centering
	\includegraphics[height=0.5\textwidth, width=1\textwidth]{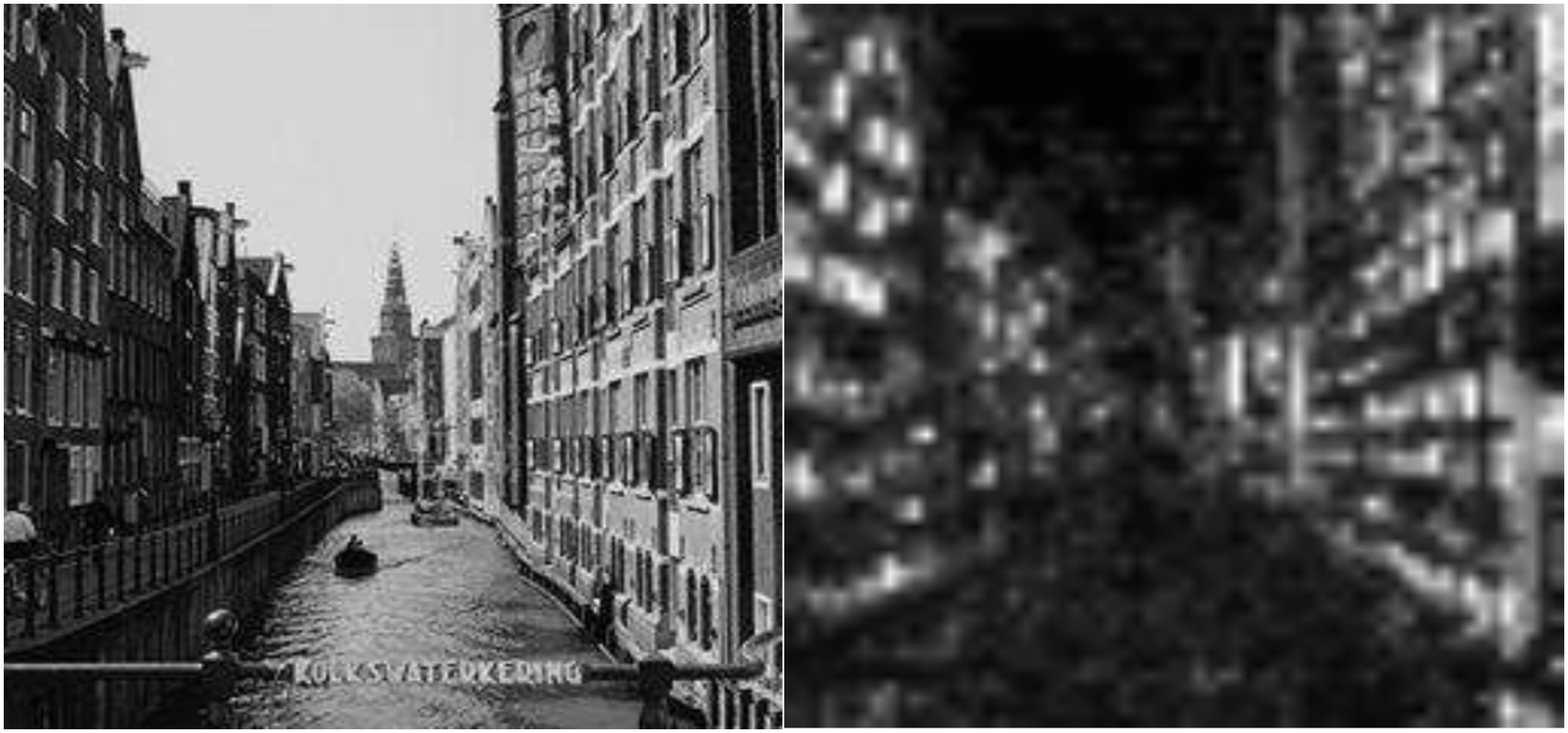} \\ 
	\caption{MIT tall building}	
\end{subfigure}
\begin{subfigure}[b]{0.15\textwidth}
	\centering
	\includegraphics[height=0.5\textwidth, width=1\textwidth]{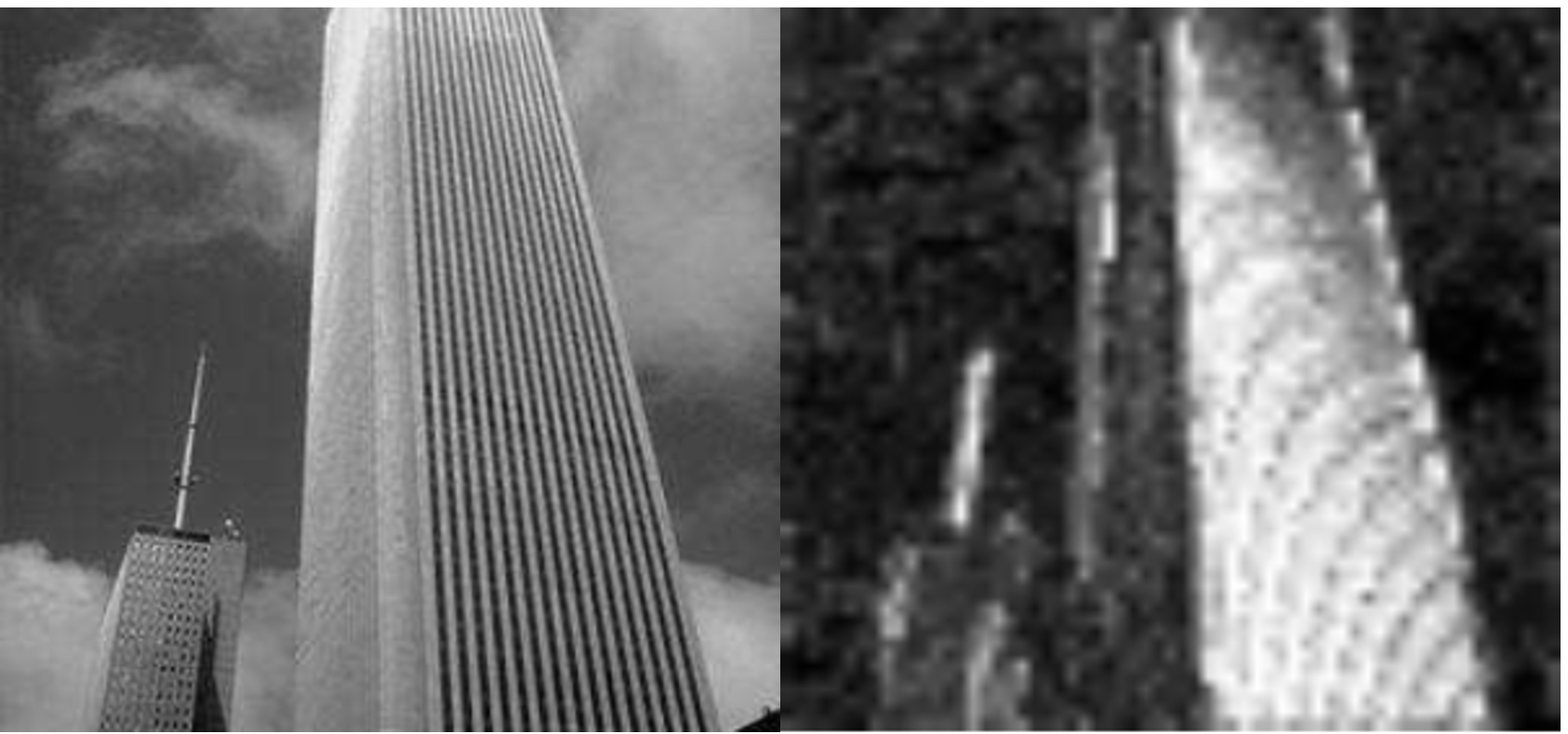} \\ 
	\caption{MIT street}	
\end{subfigure}
\begin{subfigure}[b]{0.15\textwidth}
	\centering
	\includegraphics[height=0.5\textwidth, width=1\textwidth]{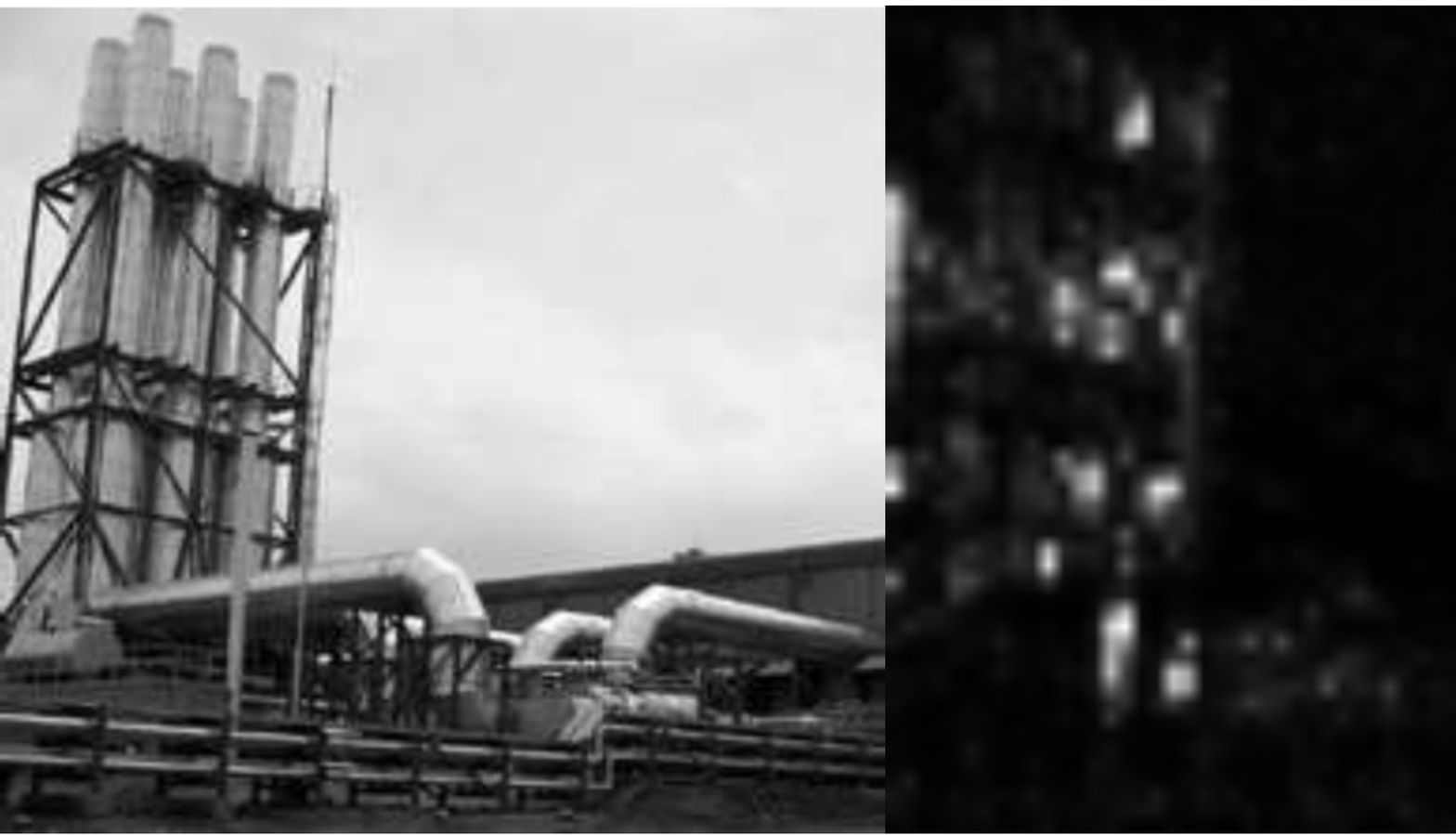} \\ 
	\caption{Industrial}	
\end{subfigure}
\begin{subfigure}[b]{0.15\textwidth}
	\centering
	\includegraphics[height=0.5\textwidth, width=1\textwidth]{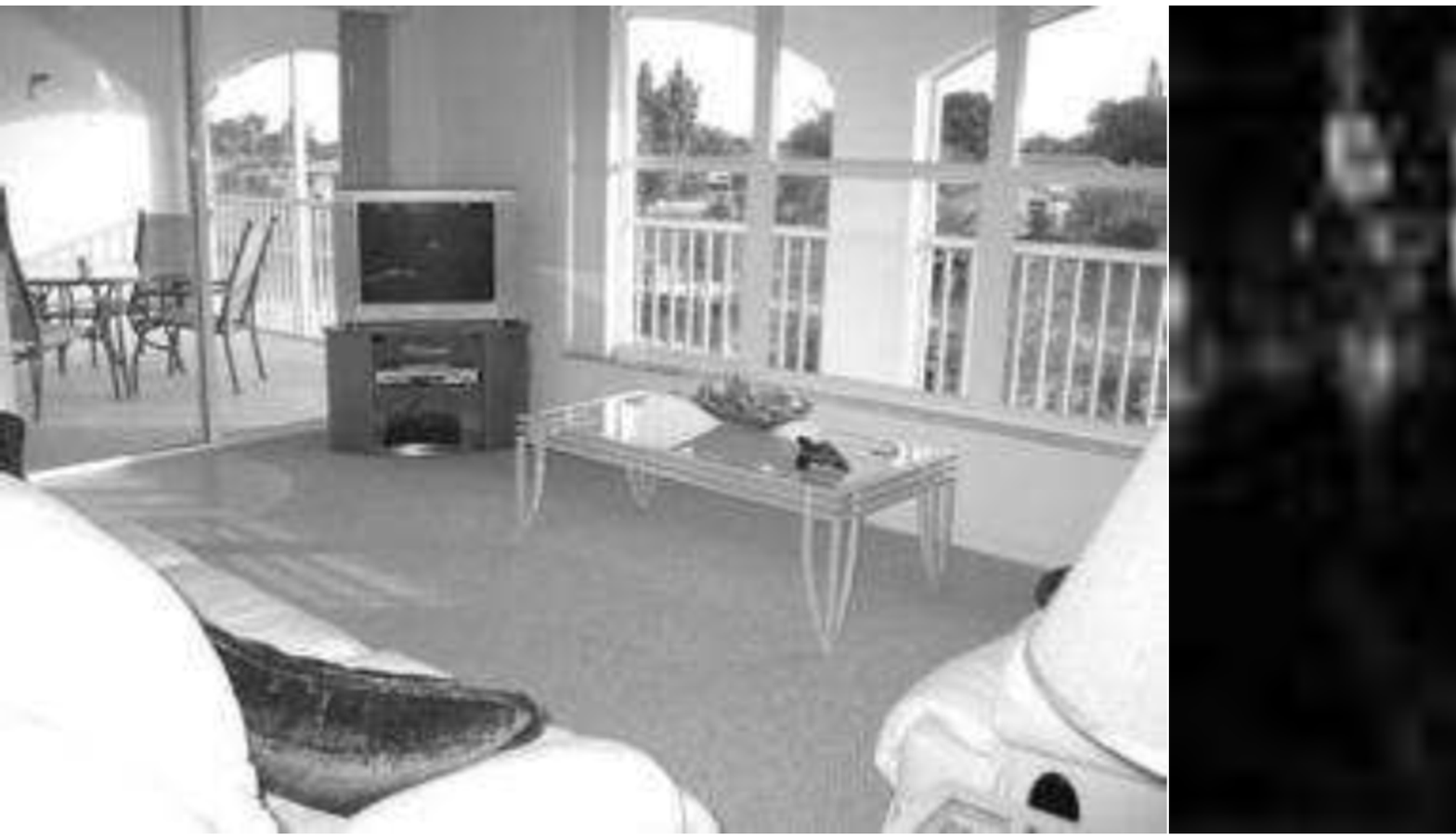} \\ 
	\caption{Living room}	
\end{subfigure}
\caption{$p(c|x,d)$ visualization (right column) on selected images (left column) in the 15-Scene dataset. Best viewed in color.}
\label{fig:Pc_xd_scene15}
\end{figure}

%=============================================================================================

%============================================================================
\section{Discussion}
\label{r:Sec5}
\begin{figure}[t]
\centering
\begin{subfigure}[b]{0.15\textwidth}
	\centering
	\includegraphics[height=0.5\textwidth, width=1\textwidth]{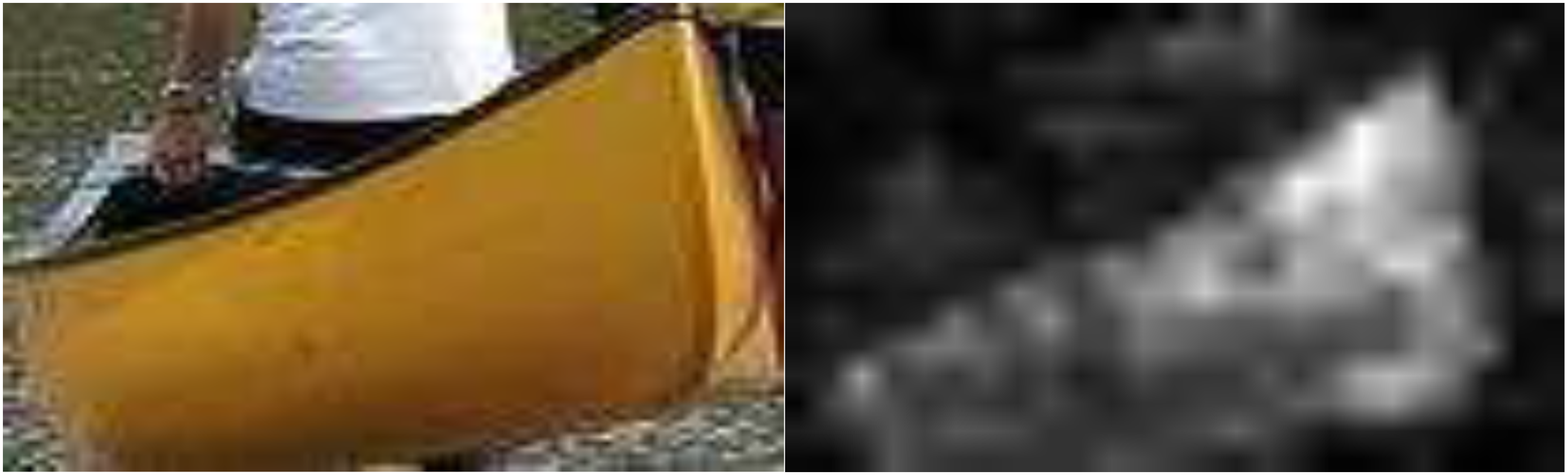} \\ 
	\caption{Boat}	
\end{subfigure}
\begin{subfigure}[b]{0.15\textwidth}
	\centering
	\includegraphics[height=0.5\textwidth, width=1\textwidth]{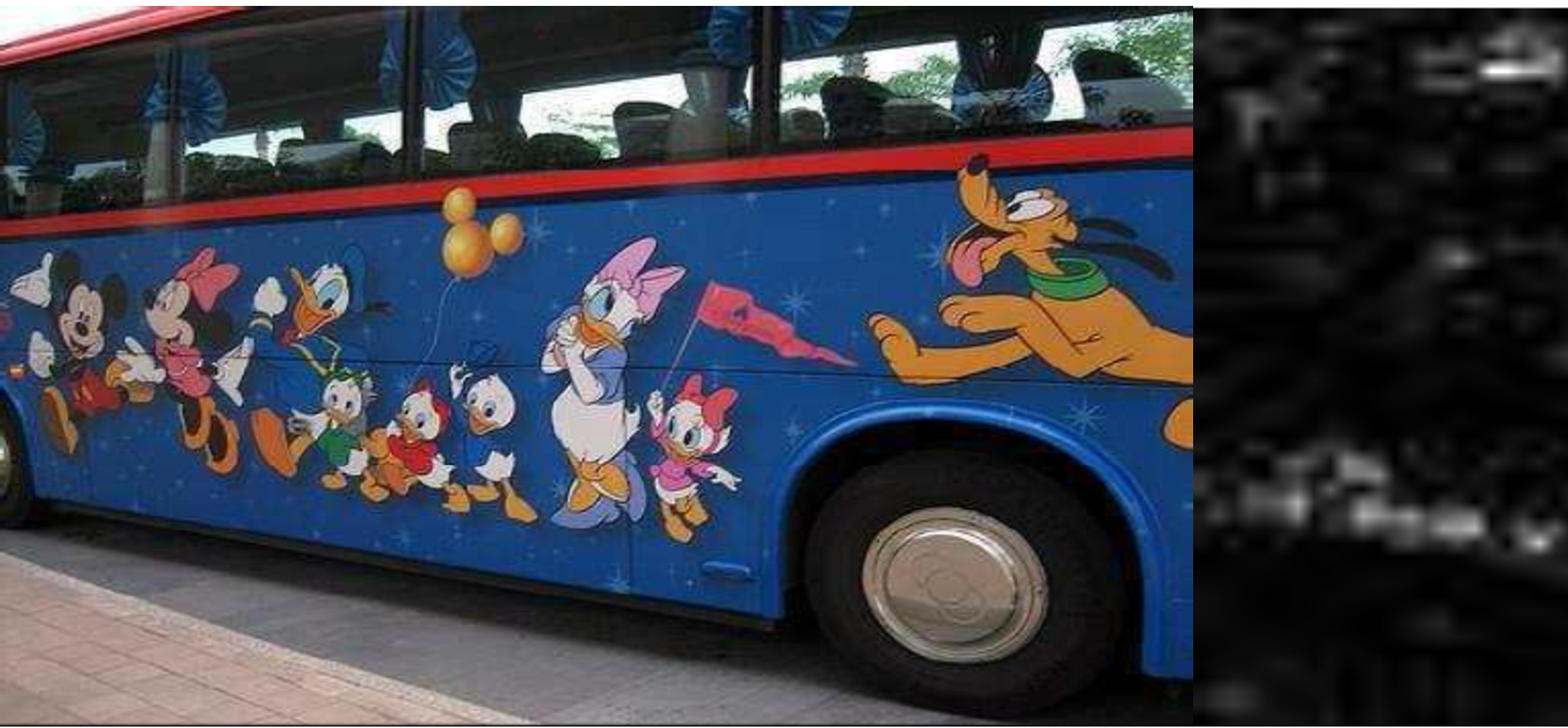} \\ 
	\caption{Bus}	
\end{subfigure}
\begin{subfigure}[b]{0.15\textwidth}
	\centering
	\includegraphics[height=0.5\textwidth, width=1\textwidth]{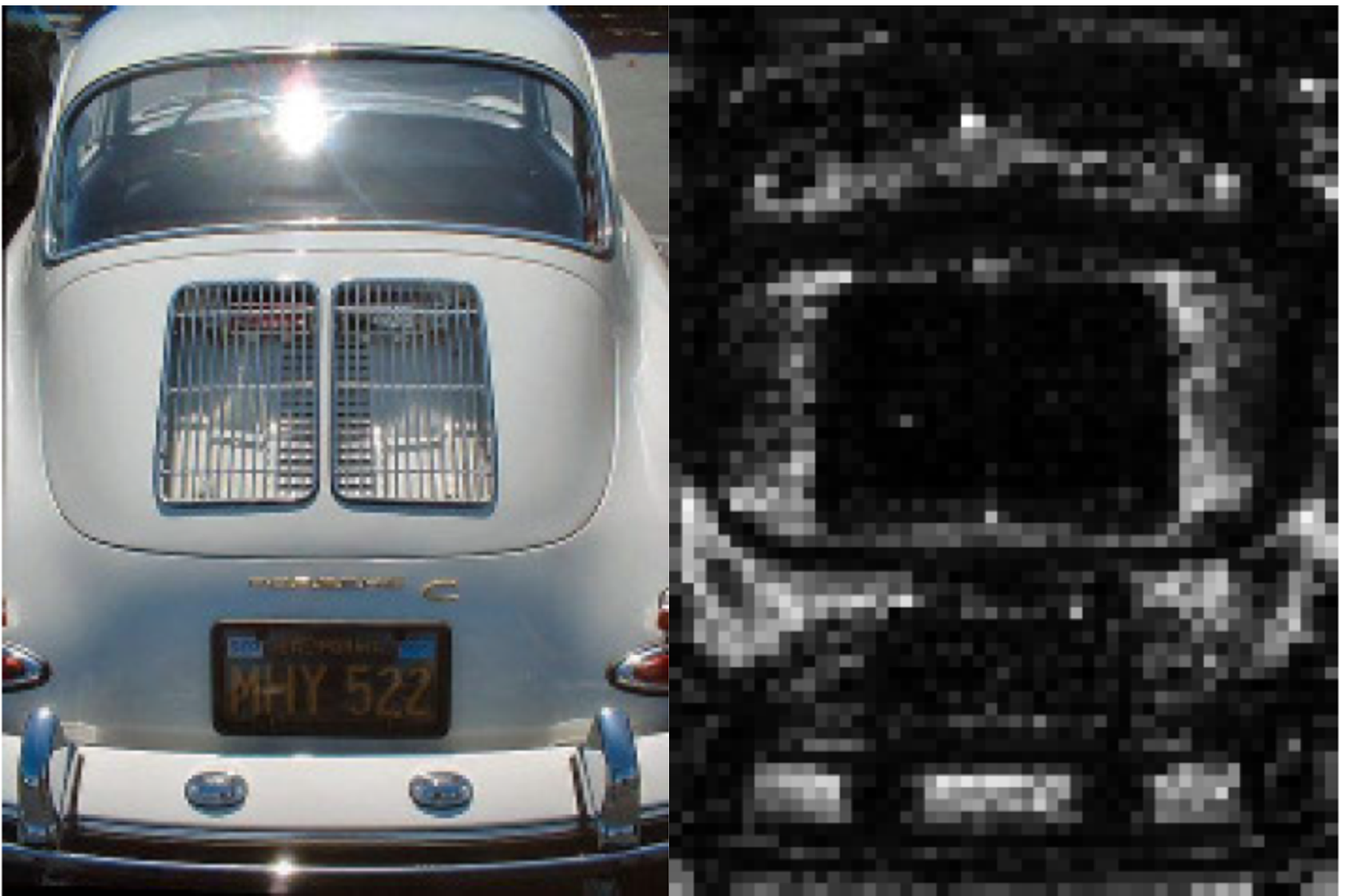} \\ 
	\caption{Car}	
\end{subfigure}

\begin{subfigure}[b]{0.15\textwidth}
	\centering
	\includegraphics[height=0.5\textwidth, width=1\textwidth]{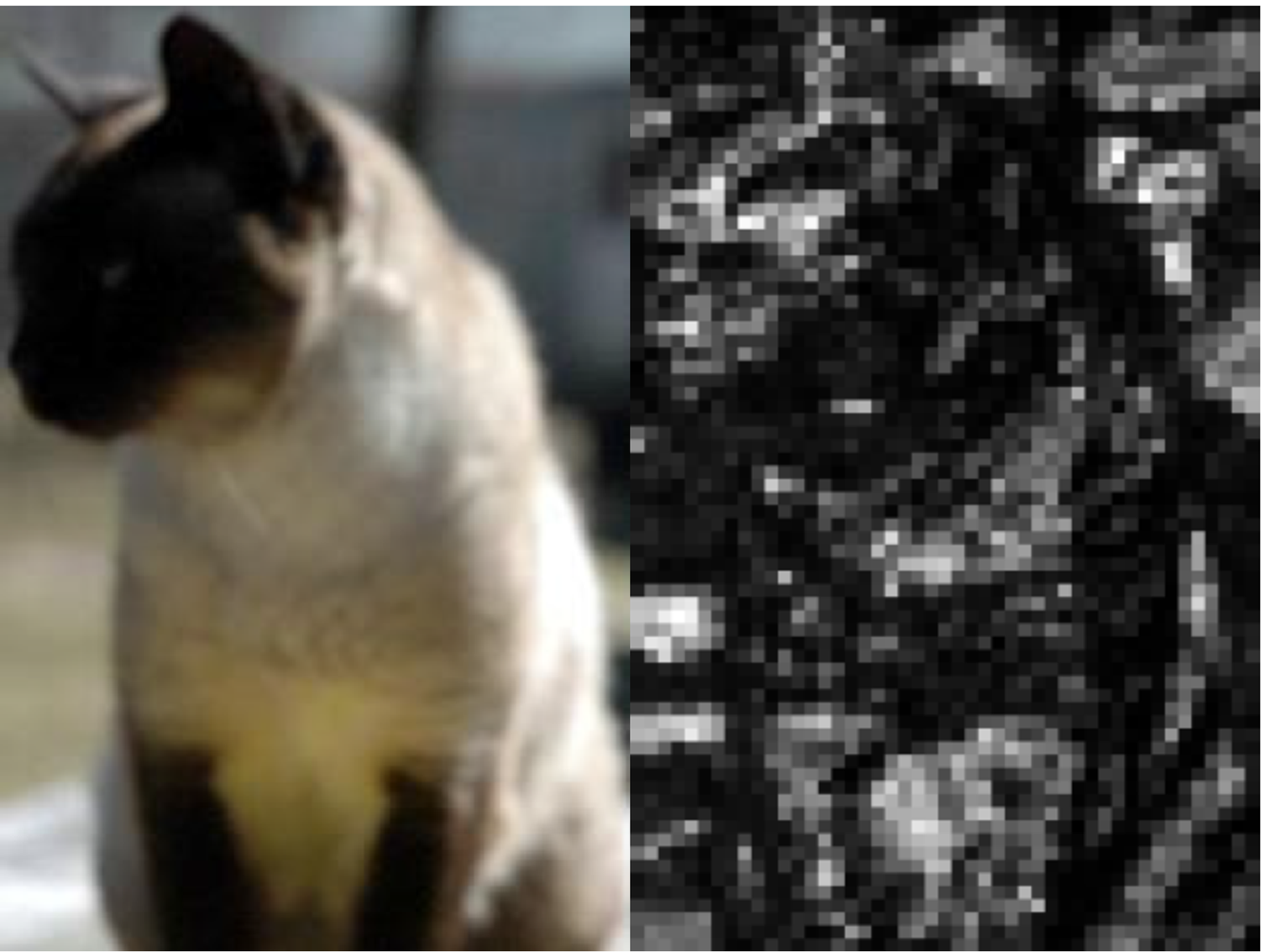} \\ 
	\caption{Cat}	
\end{subfigure}
\begin{subfigure}[b]{0.15\textwidth}
	\centering
	\includegraphics[height=0.5\textwidth, width=1\textwidth]{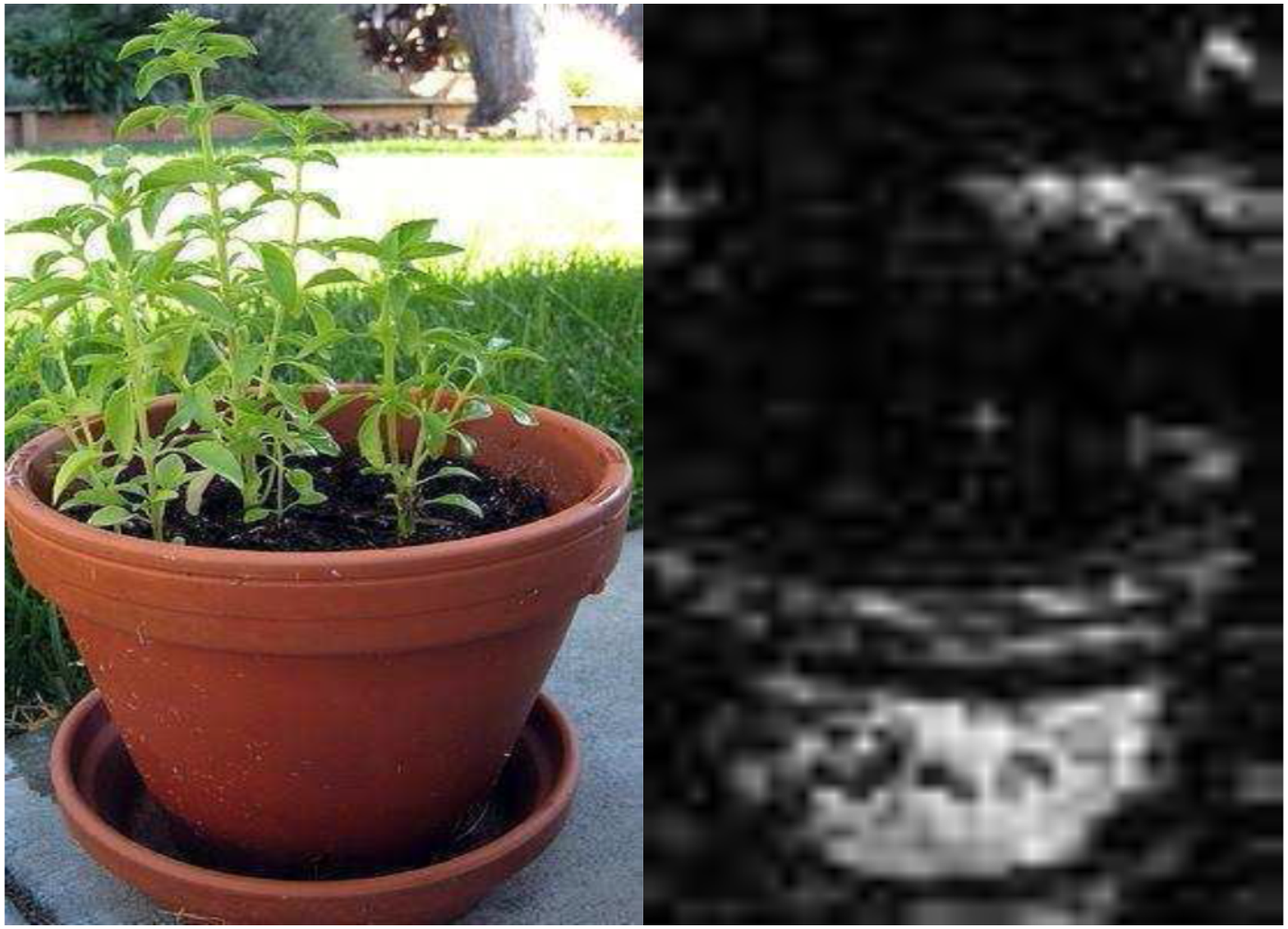} \\ 
	\caption{Potted plant}	
\end{subfigure}
\begin{subfigure}[b]{0.15\textwidth}
	\centering
	\includegraphics[height=0.5\textwidth, width=1\textwidth]{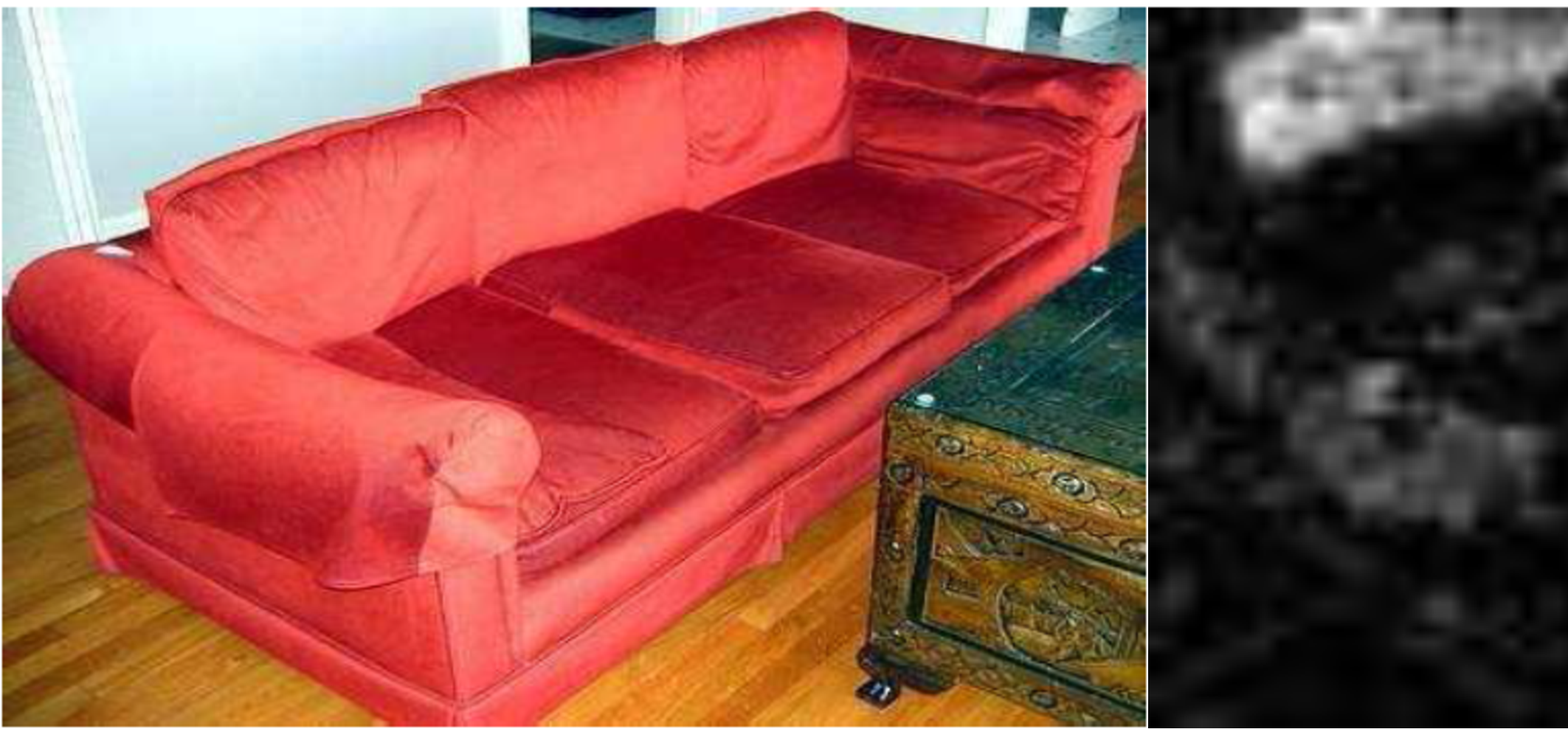} \\ 
	\caption{Sofa}	
\end{subfigure}
\caption{$p(c|x,d)$ visualization (right column) on selected images (left column) in the C-Pascal dataset. Best viewed in color. }
\label{fig:Pc_xd_CPascal}
\end{figure}

For the proposed method to work effectively, the soft class labels play a major role. In here, we visualize the image-patch-specific class distribution $p(c|x,d)$ to review the effects of soft class labels during the codebook updating process. 
Visualization of $p(c|x,d)$ for the 15-Scene and C-Pascal datasets are illustrated in Figure \ref{fig:Pc_xd_scene15}-\ref{fig:Pc_xd_CPascal} respectively. We show that $p(c|x,d)$ represents a rough silhouette to the original image. 
Besides, the high probability area (white area) normally reflects the edges of the image, that is reflects the characteristic of the images especially objects in the image. $p(c|x,d)$ can be considered as an error reduction in the RF learning. By assigning background patches as low probability area, we reduce the probability that the background patches are employed in the RF node splitting, and hence improve the RF discriminative power.

On another aspect, the computational cost of the proposed method depends on the number of iterations as one iteration consists of RF codebook learning and pLSA learning. However, subsequent iterations get accelerated as we just repeat the learning process by using soft class labels instead of ordinary class labels.
%****************************************************************************

%============================================================================
% Conclusion and Future Work
%============================================================================
\section{Conclusion}
\label{r:Sec6}
In this paper, we proposed a novel feedback framework which utilizes the discriminative RF codebook learning and generative classifier learning in image understanding task. To achieve that, we estimate soft class labels from the initial pLSA model and RF codebook to update the RF codebook iteratively until convergence is reached. We show that this framework can be applied in SSL paradigm as well.
The future work is to investigate different feature extraction parameters effect (e.g. patch size and step size) on the soft class labels learning. Besides, we are also interested to find a more robust way for the convergence decision. 

%****************************************************************************
\section*{Acknowledgment}
This research is supported by the University Malaya Research Grant (UMRG) Grant RP023-2012D, H-000000-56657-E13110 and the Bright Sparks Programme (BSP) from the University of Malaya.




%\bibliographystyle{IEEEtran}
%\begin{thebibliography}{1}
%\end{thebibliography}

{\small
\bibliographystyle{ieee}
\bibliography{egbib}
}

\end{document}